\documentclass[10pt,twocolumn,letterpaper]{article}

\usepackage[pagenumbers]{ijcai} 

\usepackage{multirow}
\usepackage{bbding}
\usepackage{newtxmath}
\usepackage{algorithm}
\usepackage{algorithmic}

\def\etal{{\em et al.\/}\, }

\def\etal{{\em et al.\/}\, }

\makeatletter
\DeclareRobustCommand\onedot{\futurelet\@let@token\@onedot}
\def\@onedot{\ifx\@let@token.\else.\null\fi\xspace}

\def\eg{\emph{e.g}\onedot} 
\def\ie{\emph{i.e}\onedot} 
 
 \def\vs{\emph{vs}\onedot}

\def\etal{\emph{et al}\onedot}
\makeatother

\def\0{{\bf 0}}
\def\1{{\bf 1}}

\def\bd{{\bf d}}

\def\bo{{\bf o}}

\def\br{{\bf r}}

\def\bx{{\bf x}}

\def\citep{\cite}
\def\citet{\cite}

\newcommand{\tablestyle}[2]{\setlength{\tabcolsep}{#1}\renewcommand{\arraystretch}{#2}\centering\footnotesize}
\newlength\savewidth

\usepackage{times}
\usepackage{epsfig}
\usepackage{graphicx}
\usepackage{amsmath}
\usepackage{amssymb}
\usepackage{booktabs}
\usepackage{multirow}
\usepackage{tabularx}
\usepackage{enumitem}
\usepackage{gensymb}
\usepackage{colortbl}
\usepackage{caption}
\usepackage{makecell}
\usepackage{wrapfig}
\usepackage{amssymb}
\usepackage{pifont}

\usepackage[normalem]{ulem} 

\definecolor{mycyan}{cmyk}{.1,0,0,0}
\newcommand{\cmark}{\ding{51}}%
\definecolor{mygray}{gray}{.95}

\newcommand\mypara[1]{\vspace{2mm}\noindent\textbf{#1}}

\definecolor{ijcaiblue}{rgb}{0.21,0.49,0.74}
\definecolor{gtdepth}{rgb}{1,0.8,1}
\definecolor{metricdepth}{rgb}{0.4,0.8,1}
\definecolor{reldepth}{rgb}{0.4,1,0.8}
\usepackage[pagebackref,breaklinks,colorlinks,allcolors=ijcaiblue]{hyperref}

\title{Learning A Zero-shot Occupancy Network from Vision Foundation Models via Self-supervised Adaptation}

\author{
Sihao Lin\quad Daqi Liu\quad Ruochong Fu\quad Dongrui Liu\quad Andy Song\quad Hongwei Xie\quad \\ Zhihui Li \quad Bing Wang\quad Xiaojun Chang \\
{\normalsize
RMIT University\; Xiaomi Auto\; SJTU\; USTC\; UTS}\\
}

\begin{document}
\maketitle
\begin{abstract}
Estimating the 3D world from 2D monocular images is a fundamental yet challenging task due to the labour-intensive nature of 3D annotations. To simplify label acquisition, this work proposes a novel approach that bridges 2D vision foundation models (VFMs) with 3D tasks by decoupling 3D supervision into an ensemble of image-level primitives, e.g., semantic and geometric components. As a key motivator, we leverage the zero-shot capabilities of vision-language models for image semantics. However, due to the notorious ill-posed problem - multiple distinct 3D scenes can produce identical 2D projections, directly inferring metric depth from a monocular image in a zero-shot manner is unsuitable. In contrast, 2D VFMs provide promising sources of relative depth, which theoretically aligns with metric depth when properly scaled and offset. Thus, we adapt the relative depth derived from VFMs into metric depth by optimising the scale and offset using temporal consistency, also known as novel view synthesis, without access to ground-truth metric depth. Consequently, we project the semantics into 3D space using the reconstructed metric depth, thereby providing 3D supervision. Extensive experiments on nuScenes and SemanticKITTI demonstrate the effectiveness of our framework. For instance, the proposed method surpasses the current state-of-the-art by 3.34\% mIoU on nuScenes for voxel occupancy prediction.
\end{abstract}
\section{Introduction}
\label{sec:intro}
The vision-centric paradigm~\cite{ma2024vision,izadinia2017im2cad,saxena2005learning} aims to understand the 3D environment using RGB images. It has witnessed great success in autonomous driving (AD) for versatile applications such as object detection~\cite{huang2021bevdet,wang2021progressive} and occupancy prediction~\cite{vobecky2024pop,huang2024selfocc}. Nonetheless, the labour-intensive annotation process~\cite{behley2019semantickitti,caesar2020nuscenes,tian2024occ3d} has long been the Achilles' heel in vision-centric AD systems. On one hand, we aim to train a 3D driving system from effortless 2D data. However, many camera sensor-based car models can only capture plaint images but not semantic and geometric priors, posing challenges to 3D annotations. To address this issue, this work strives to build a vision-centric 3D occupancy network using an image-only dataset.

\begin{figure}
    \centering
    \includegraphics[width=0.7\linewidth]{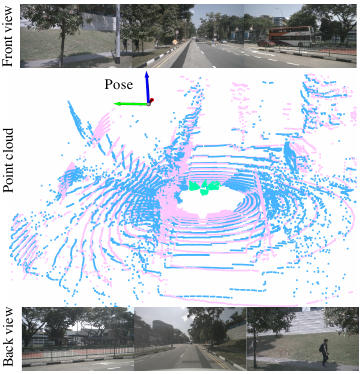}
    \vspace{-3mm}
    \caption{
    \textbf{Adapting relative depth into metric depth}. Existing VFM~\cite{yang2024depth} delivers promising relative depth while less capable of metric depth due to ill-posedness. Without access to ground truth, our method leverages the novel view synthesis (\cref{sec:scale}) to calibrate the relative depth \fcolorbox{white}{reldepth}{\rule{0pt}{3pt}\rule{3pt}{0pt}} (zoom out if necessary) into metric one \fcolorbox{white}{metricdepth}{\rule{0pt}{3pt}\rule{3pt}{0pt}}, which aligns well with ground truth depth \fcolorbox{white}{gtdepth}{\rule{0pt}{3pt}\rule{3pt}{0pt}}.
     }
    \label{fig:shoutu}
\vspace{-3mm}
\end{figure}

As a motivator, one may resort to 3D vision foundation models (VFMs)~\cite{zhou2023uni3d,xu2023pointllm,yang2023lidar,hong20233d,chen2023ll3da} to facilitate 3D annotations. Unfortunately, existing literature on 3D VFMs necessitates 3D input (\eg, point cloud) and deviates from the data sources of vision-centric systems. On the other hand, training a vision-centric variant of 3D VFMs from scratch leads to a chicken-and-egg problem. In parallel, the emergence of 2D VFMs, such as SAM~\cite{kirillov2023segment} and DINOv2~\cite{oquab2023dinov2}, has empowered a series of 2D vision tasks~\cite{minderer2024scaling,li2022languagedriven,zhang2023clip,peng2023openscene,kerr2023lerf}, demonstrating a great ability to ease the acquisition of image-level primitives. Therefore, it is tempting to exploit the zero-shot image primitives derived from \emph{ready} 2D VFMs for 3D tasks, which remains largely untapped.

By definition, the spatial grids geometrically correspond to the image patches, which can be formulated by pixel-wise depth and camera pose. We therefore motivate connecting the 2D VFMs with 3D tasks by decoupling the 3D supervision signals into image-level semantics and geometry independently. Specifically, given an RGB image dataset without any semantic or geometric labels, we apply vision foundation models on the monocular image to obtain the pixel-wise semantic and depth primitives separately. Accordingly, we project the semantics into the 3D space using the depth information and camera parameters, yielding an alternative to 3D supervision (See Fig.~\ref{fig:pixel2voxel}).

Despite decade-long research~\cite{saxena2005learning,wei2023surrounddepth,yang2024depth}, current depth estimation methods are less capable of \textbf{\emph{metric}} depth in unseen domains, primarily due to the notorious ill-posedness. That is, different 3D scenes can lead to identical 2D projections. Thus, two images that appear identical may have different depth information, making it challenging to zero-shot metric depth.  While some efforts~\cite{guizilini2023towards,yin2023metric3d,radford2018improving} have been devoted to zero-shot metric depth, we observe they are less capable of occupancy prediction. Existing VFMs~\cite{yang2024depth,depth_anything_v2} offer appealing sources of \textbf{\emph{relative}} depth in a zero-shot fashion, which can be calibrated to metric depth with appropriate scale and offset. Inspired by this, we propose to prompt the relative depth derived from VFMs into metric depth by optimising the scale and offset without access to ground-truth depth. Specifically, we leverage novel view synthesis (NVS) among video sequences, optimise the scale and offset to minimise the reconstruction error (See~\cref{fig:nvs}). Zhang~\etal~\cite{zhang2022hierarchical} noted that recovering metric depth requires different scales and offsets for each image pixel, making optimisation exponentially difficult. Further, we found that naively applying gradient descent~\cite{Kingma2014AdamAM,Loshchilov2019DecoupledWD} leads to divergence due to sensitive initialisation.

To this end, we develop a two-stage coarse-to-fine optimisation strategy. In the first stage, we traverse an array of candidate scales (single scalar) applied to the whole scene and select the one with minimum reconstruction error. In the next stage, we initialise pixel-wise scales with the holistic scene scale from the previous stage and optimise them, together with a scene offset, via gradient descent.

We evaluate the proposed framework for 3D occupancy prediction on nuScenes~\cite{caesar2020nuscenes} and SemanticKITTI~\cite{behley2019semantickitti}, which requires classifying the voxel semantic of the 3D volume in the driving environment. Occupancy prediction is an important task for autonomous driving because it can serve as a proxy for scene understanding given the granularity.
Our framework can surpass, sometimes strikingly, the existing methods~\cite{huang2024selfocc,cao2023scenerf} with bird's eye view (BEV) representation. For instance, our method outperforms SelfOcc~\cite{huang2024selfocc} by 3.34\% mIoU in nuScenes. Our work unveils another direction for vision-centric 3D tasks by exploiting the 2D VFMs. We hope our findings can inspire other dimensions of 3D visions. To summarise, the contributions of this work are:

\begin{itemize}
    \item 
    We present the first vision-centric framework to connect the image primitives derived from 2D VFMs with 3D tasks, enabling zero-shot learning for 3D occupancy prediction with effortless label-free images.
    
    \item We propose a two-stage coarse-to-fine optimisation process to calibrate the high-fidelity relative depth of VFM into metric depth via temporary consistency where no ground truth depth is needed.
    
    \item Extensive experiments on nuScenes and SemanticKITTI demonstrate our framework can facilitate the 3D occupancy prediction with no label available, advancing the state-of-the-art of vision-centric paradigm.
\end{itemize}
\section{Related Work}
\paragraph{Vision foundation models.}
Large models first emerged in the natural language processing field~\cite{radford2018improving,radford2019language,brown2020language,touvron2023llama,touvron2023llama2}, and the idea was later introduced to computer vision. A wide range of recent research~\cite{wang2023tracking,oquab2023dinov2,kirillov2023segment} has focused on 2D Vision Foundation Models (VFMs), including text-to-image synthesis~\cite{saharia2022photorealistic,ruiz2023dreambooth}, open-vocabulary segmentation~\cite{peng2023openscene,li2022languagedriven}, and depth estimation~\cite{yang2024depth}. Unfortunately, 3D VFMs are challenging to develop following the pretrain-finetune paradigm. This is because 3D data (annotations) are difficult to collect compared to the vast number of high-quality images, and the hardware configurations are not standardised, \eg, beam number and emission rate of LiDAR sensors. Furthermore, existing 3D VFMs~\cite{zhou2023uni3d,xu2023pointllm,yang2023lidar,hong20233d,chen2023ll3da} require 3D input, \eg, point cloud, and are not well suited to the vision-centric paradigm, where the aim is to understand the 3D world using 2D images. 

\mypara{Distilling 2D VFM for AD scenario.}
Training with extensive high-fidelity 2D data, it is tempting to leverage 2D VFMs to enhance AD tasks by serving as auxiliary supervision signals. A promising approach~\cite{vobecky2024pop,liu2024segment,chen2024towards} involves distilling the learned representation from 2D VFMs into 3D models. Unfortunately, this strategy has significant limitations, as it requires the mandatory use of LiDAR sensors for 3D prior (\ie, point cloud) during both training and inference stages. In contrast, our work aims to develop 3D occupancy prediction using unlabelled images without 3D sensors, enabling wider application scenarios.

\mypara{Zero-shot semantic and metric depth.}
CLIP~\cite{radford2021learning} is a two-tower structure in which the image features are aligned with language descriptions, allowing zero-shot image classification. The image-level alignment has lately been augmented to pixel-level alignment~\cite{kerr2023lerf,zhang2023clip,li2022languagedriven}, empowering zero-shot pixel-wise semantic. Regarding metric depth, it has long been a significant challenge for computer vision due to notorious ill-posedness. We refer the reader to Figure 3 of Metric3D~\cite{yin2023metric3d} for details of ill-posedness. Despite challenges, some works~\cite{guizilini2023towards,yin2023metric3d}  focus on zero-shot metric depth, which encodes the camera intrinsics into representation learning. Recently, DepthPro~\cite{bochkovskii2024depth} introduces an extra branch to predict the focal length without needing the camera intrinsic. Nonetheless, their zero-shot metric depths are unsatisfactory in occupancy prediction~\cite {caesar2020nuscenes} (See~\cref{sec:exp_occ}). Recent literature~\cite{yang2024depth} trains a relative depth estimator using a mixture of available benchmarks. This work uses 2D VFM’s relative depth and calibrates it to metric depth using self-supervised adaptation (\cref{sec:scale}), balancing ill-posedness and metric depth. 

\mypara{Occupancy prediction.} 
The proposed framework is evaluated on occupancy prediction~\cite{vobecky2024pop,tian2024occ3d,li2023fb,huang2023tri,zhang2023occformer}, which has received considerable attention in the field of autonomous driving due to its significance. Amongst existing literature, the vision-centric paradigm~\cite{huang2024selfocc,cao2023scenerf} attracts great interest, as cameras are easier and more cost-effective to deploy on autonomous vehicles. Despite successes, existing methods still require laborious 3D annotations. Surprisingly, this problem has only recently been studied by a small number of pioneering works~\cite{cao2023scenerf,huang2024selfocc}, which proposed acquiring geometric information through self-supervised learning. Typically, these methods learn geometric representations by optimising the temporal and multi-view consistency of sequential images using ray casting~\cite{mildenhall2021nerf}. Consequently, they implicitly encode depth information by learning to warp images across different time steps and viewpoints. 
\section{Method}
We begin by briefly introducing the preliminaries of occupancy prediction in Sec.~\ref{sec:preliminaries}. Next, we formulate the approach of decoupling 3D signals into image primitives in Sec.~\ref{sec:formulate}, facilitating the connection between 2D VFMs and 3D occupancy prediction (Sec.~\ref{sec:decouple}). In~\cref{sec:scale}, we propose a two-stage coarse-to-fine optimisation strategy to align relative depth with metric depth.

\subsection{Preliminaries}
\label{sec:preliminaries}
Given an image set $\mathcal{I}=\{\bx_{v}\}_{v=0}^{N-1}$ consists of $N$ synchronized camera views, where $\bx \in \mathbb{R}^{3\times H\times W}$ denote the RGB image of height $H$ and width $W$, the goal of occupancy prediction is to perform $K$ way classification on the voxel volume $V\in \mathbb{R}^{X\times Y \times Z}$. In the vision-centric paradigm, occupancy prediction can be formulated as:
\begin{equation}
    \hat{O} = 
    \mathop{\arg\max}\limits_{k\in [0,1,\cdots,K-1]} 
    F(\mathcal{I}|\theta)\in \mathbb{R}^{X\times Y\times Z \times C_k},
\end{equation}

\noindent where $F$ is the target network parameterised by $\theta$ and $\hat{O}$
is the per-voxel prediction (occupied status and category). In the conventional fully-supervised learning~\cite{tian2024occ3d}, $\hat{O}$ is optimised to minimise the cross-entropy with ground-truth. 
\begin{figure}[t!]
\centering
\vspace{0mm}
\includegraphics[width=0.8\linewidth]{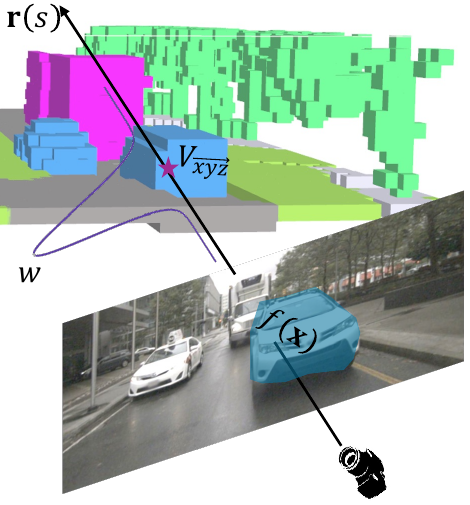}
\caption{Decoupling 3D signal as image primitives.}
\label{fig:pixel2voxel}
\vspace{0mm}
\end{figure}

\begin{figure*}[h]
    \centering
        \includegraphics[width=0.98\linewidth]{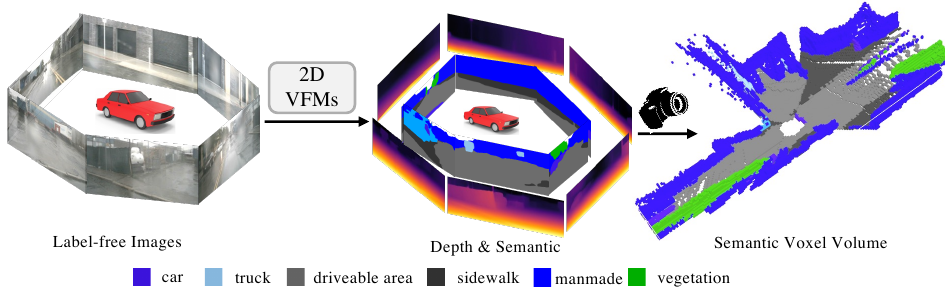}
    \caption{Scheme of the proposed method. We propose decoupling the 3D signals into image primitives, allowing the connection between 2D VFMs and 3D tasks. Given label-free images, their semantic and geometry information are derived from 2D VFMs via self-supervised adaptation (\cref{sec:decouple}). The ensemble of the image primitives serves as the alternative to 3D supervision.}
    \label{fig:framework}
\end{figure*}

\subsection{Decoupling 3D Signals As Image Primitives}
\label{sec:formulate}
Given label-free images, our objective is to obtain high-quality alternatives to 3D supervision signals. While directly deriving alternatives from a 3D model seems feasible, no effective 3D function currently addresses this. Alternatively, semi-supervised learning would restrict the category space, limiting its applicability. In contrast, this work proposes decoupling 3D signals into image primitives. For a voxel grid, its semantics can be formulated as the sum of the ray of image semantics within the space:

\begin{equation}
   \underset{\underset{\rm 3D\ sem.}{\downarrow}}{V_{ \overrightarrow{xyz}}} =
   \mathop{\arg\max}\limits_{k\in [0,1,\cdots,K-1]} 
   \sum_{v}\sum_{\br}\underset{\underset{\rm one-hot\ sem.}{\downarrow}}{f(\bx_{v})} \cdot w ,
\label{eq:decouple}
\end{equation}

\begin{equation}
w=\left\{
\begin{aligned}
    &1, \ \ || \Gamma^{-1}(d(\bx),\br) - \overrightarrow{xyz} || < \epsilon, \\
    &0, \ \ {\rm otherwise},
\end{aligned}
\right.
\label{eq:weight}
\end{equation}

\begin{equation}
\begin{aligned}
\br(s) &= \bo + s\cdot \bd,
\end{aligned}
\label{eq:ray}
\end{equation}

\noindent where $V_{ \overrightarrow{xyz}}$ denotes the resulting 3D semantic at the coordinate $[x,y,z]$ along the ray $\br$ across camera views $v$ , where $\bo$ is the camera origin, $\bd$ represents the direction and $s$ is the step. We omit the symbol $s$ as we consider all the steps in the space. The term $f(\bx) \in \mathbb{R}^{H \times W \times C_k}$ is the one-hot semantic vector, and $w$ is the Dirac distribution that decides whether to assign image semantics at $[x,y,z]$. Accordingly, we define $w$ as a depth estimation function $d(\cdot)$ in~\cref{eq:weight}. Given the estimated depth $d(\bx)$, we back-project the image pixel using the projection matrix $\Gamma$. Specifically, if $\Gamma^{-1}(d(\bx), \br)$ approaches the voxel $[x,y,z]$ within a margin of error $\epsilon$, the image semantic is assigned to this location.

As illustrated in Fig.~\ref{fig:pixel2voxel}, the image semantic (cyan mask) is transmitted along a ray, reaching a voxel grid indicated by the purple star, where $w$ allocates the majority of the weight. Due to occlusion, the image semantic is not assigned to subsequent voxels, as $w$ sharply decreases. Thus, Eq.~\ref{eq:decouple} expresses the 3D representation as a series sum of the ray of 2D primitives, \ie, semantic $f$ and depth $d$, which are independent distributions. We thereby propose a zero-shot, vision-centric paradigm that decouples 3D signals into image-level semantic and depth information, leveraging the knowledge embedded in 2D VFMs.

\subsection{Zero-shot 3D Occupancy Prediction}
\label{sec:decouple}
In this section, we detail the process of extracting image primitives from the 2D VFMs. In particular, we propose a two-stage coarse-to-fine optimisation approach to calibrate the appealing relative depth into metric depth.

\subsubsection{Image Semantic Generation} 
Traditional semantic segmentation approaches are constrained to a predefined category space~\cite{zhou2017scene,lin2014microsoft}, limiting their flexibility for AD systems when addressing diverse scene distributions. To overcome this limitation, we leverage the zero-shot capability of pre-trained vision-language models~\cite{radford2021learning,li2022languagedriven} as semantic sources across various scenarios. Assume that we are interested in the category space $\mathcal{C} = \{c_k\}_{k=0}^{K-1}$, with the corresponding text embeddings represented by $\mathcal{E} = \{e_{c_k}\}_{k=0}^{K-1}$. The semantic label of each pixel is determined by calculating the cosine similarity between the text embedding and pixel feature:

\begin{equation}
\begin{aligned}
e_{c_i} = f_{\rm text}(c_k), \ k&=0,1,\cdots,K-1 ,\\
f(\bx) = \mathop{\arg\max}\limits_{c_k} f_{\rm image}(\bx) &\otimes [e_{c_{\varnothing}}, e_{c_0}, \cdots, e_{c_{K-1}}]^{\top},
\label{eq:sem}
\end{aligned}
\end{equation}

\noindent where $f_{\rm image}$ and $f_{\rm text}$ denote the image encoder and text encoder, respectively. $f_{\rm image}(\bx)$ representing the pixel-wise feature of the image $\bx$. For notation simplicity, we omit the symbol for camera view. The category ${c_{\varnothing}}$ indicates classes outside of $\mathcal{C}$, such as the sky. To obtain the text embeddings $\mathcal{E}$, we map each category name into the vocabulary table and query the pretrained CLIP text encoder, $f_{\rm text}$~\cite{radford2021learning}. We use the image encoder from LSeg~\cite{li2022languagedriven} as our implementation of $f_{\rm image}$. Pixels identified as belonging to category ${c_{\varnothing}}$ are ignored during training. 

\subsubsection{Calibrating Relative Depth into Metric Depth}
\label{sec:scale}
\paragraph{Novel view synthesis as self-supervised objective.}
As discussed in~\cref{sec:formulate}, another critical component in establishing the connection between a pixel and a voxel is the pixel-wise depth, \ie, the function $d(\cdot)$ in~\cref{eq:weight}. Although recent advancements provide promising sources of relative depth, directly inferring metric depth from VFMs remains challenging, primarily due to ill-posedness~\cite{yin2023metric3d}. Rather than fine-tuning the VFM for metric depth, this work proposes adapting relative depth to metric depth through novel view synthesis~\cite{godard2019digging}. 

\begin{figure}
    \centering
    \includegraphics[width=0.8\linewidth]{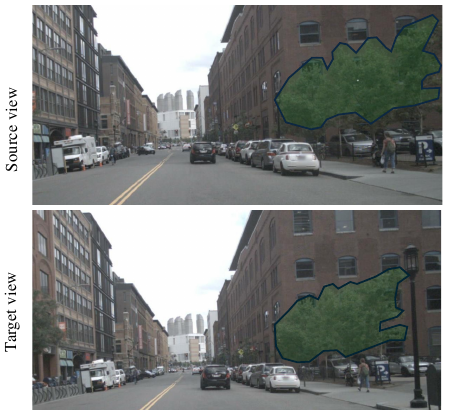}
    \vspace{-3mm}
    \caption{Illustration of novel view synthesis. We aim to reconstruct the target view from the source view by identifying the pixel correspondence between two views (\eg, the tree indicated by green mask) by~\cref{eq:ttos}. Consequently, we can optimise the depth scale by minimising the photometric loss~\cref{eq:totalloss}.}
    \label{fig:nvs}
\end{figure}

Specifically, given a target view $\bx_{t}$, we aim to \emph{reconstruct} it from the source view $\bx_{s}$ with a different camera pose within a video sequence, as illustrated in~\cref{fig:nvs}. Let $p_t$ denote the homogeneous coordinates of the pixel in the target view, with its depth represented by $d({p_t})$. We can project $p_t$ to the source view as follows:
\begin{equation}
    p_s = K T^{-1}_s T_t d({p_t}) K^{-1} p_t,
\label{eq:ttos}
\end{equation}

where $T_s$ and $T_t$ represent the camera poses for the source and target views, respectively, and $K$ denotes the camera intrinsics. Consequently, we can optimise $d(p_t)$ by minimising the following reconstruction loss between the target and source views:
\begin{equation}
    L_{rec} = \sum_{p_t} |\bx_{t}(p_t) - \bx_{s}(p_s) | .
\label{eq:rec}
\end{equation}

Note that $p_s$ may take continuous values. In such cases, we apply bilinear interpolation to retrieve the pixel value from the source view, ensuring differentiability and enabling end-to-end optimisation.

\mypara{Coarse-to-fine optimisation.}
Current depth VFM~\cite{yang2024depth} learns relative depth within a scene. Consequently, adapting this to metric depth becomes feasible with appropriate scaling and offset adjustments. In this work, we implement monocular depth $d(\bx) \in \mathbb{R}^{H \times W}$ as the relative depth $d_{\rm vfm}(\bx)$ from the VFM, using a scale factor $\lambda$ and offset $\gamma$:
\begin{equation}
    d(\bx) = \lambda \cdot d_{\rm vfm}(\bx) + \gamma.
\label{eq:scale}
\end{equation}

To reconstruct metric depth, \cite{zhang2022hierarchical} suggested that each pixel should be scaled independently. Consequently, $\lambda$, rather than being a scalar, shares the same resolution as the relative depth map, which introduces optimisation challenges (\cref{eq:rec}). While gradient descent might seem a viable solution, our empirical study reveals that straightforward application of gradient descent can lead to divergence. This issue arises because the reconstruction loss is sensitive to the initialisation of $\lambda$. 
To address this, we propose a coarse-to-fine strategy to optimise both $\lambda \in \mathbb{R}^{H \times W}$ and $\gamma \in \mathbb{R}$.

\begin{algorithm}[t!]
\caption{Holistic Scene Scale}
\begin{algorithmic}[1]
\STATE \textbf{Input:} camera intrinsic $K$, camera poses of source and target view $T_s$ and $T_t$, source and target view images $\bx_s$ and $\bx_t$, pixel coordinate set $\mathcal{P}$, depth VFM $d_{\rm vfm}$.
\STATE \textbf{Output:} Scene scale $\hat{\lambda}$
\STATE lossArray $\leftarrow \emptyset$
\FOR{$s = 1$ \textbf{to} $100$}
\STATE rec $\leftarrow$ 0
\FOR{$p_t$ in $\mathcal{P}$}
    \STATE $d(p_t) = sd_{\rm vfm}(p_t)$ $\triangleleft$ scaling relative depth
    \STATE $p_s \leftarrow K T^{-1}_s T_t d({p_t}) K^{-1} p_t$  $\triangleleft$ view synthesis
    \STATE rec $\leftarrow$ rec + $|\bx_{t}(p_t) - \bx_{s}(p_s)|$  $\triangleleft$ reconstruction loss
\ENDFOR
\STATE Append rec to lossArray
\ENDFOR
\STATE $\hat{\lambda} \leftarrow \mathop{\arg\min}\limits_{i}$ lossArray $\triangleleft$ scene scale
\end{algorithmic}
\label{alg:scene_scale}
\end{algorithm}

In the first stage, we search for a single scalar scale that minimises reconstruction loss at the scene level. As summarised in~\cref{alg:scene_scale}, we traverse a range of candidate scales (from 1 to 100) and compute the reconstruction error with respect to~\cref{eq:rec}. We then select the scale with the minimum loss as the scene scale, $\hat{\lambda}$.

We use $\hat{\lambda}$ from the previous stage to initialise $\lambda$. To achieve fine-grained depth, we further apply gradient descent to fine-tune the pixel-wise scale $\lambda$ alongside the scene offset $\gamma$. In addition to the reconstruction error, we incorporate SSIM~\cite{wang2004image} as an additional supervisory signal, \ie
\begin{equation}
    L_{ssim} = 0- {\rm SSIM}(\bx_t,\bx^{'}_t),
\end{equation}
\noindent where $\bx^{'}_t$ denotes the projected view of $\bx_t$ as defined in~\cref{eq:ttos}.
Finally, the objective for fine-tuning the pixel-wise scale and scene offset is given by:
\begin{equation}
    L_{total} = 0.5\cdot L_{rec} + 0.5\cdot L_{ssim}.
\label{eq:totalloss}
\end{equation}

\mypara{Masking moving objects.}
The above discussion on the metric depth pertains to static objects. 
Following~\cite{godard2019digging}, we enhance temporal consistency by masking out moving objects. In the absence of ground-truth semantics, we employ CLIP~\cite{brown2020language,li2022languagedriven} to identify image pixels corresponding to moving objects (\eg, vehicles) using~\cref{eq:sem} and exclude them from the photometric loss. The consequent scale and offset would apply to the relative depth of these moving objects.

\mypara{Method settings.} We optimise the scale and offset for each image separately to reconstruct the metric depth. The offline dataset is then used to train an occupancy prediction network. Notably, the camera poses and intrinsics (\cref{eq:ttos}) can be obtained online once the cameras are calibrated.

\begin{table*}[!t]
    \caption{Performance in nuScenes validation set.  We compare the proposed method to current state-of-the-art methods with different supervision signals.
    We report the mIoU (\%) as the evaluation metric. 3D: 3D supervision. L: LiDAR. C: Camera.
    }
    \vspace{-3mm}
    \centering
    \resizebox{2.0\columnwidth}{!}
    {
    \tablestyle{1.5pt}{1.05}
    \begin{tabular}{@{}l|c|cccccccccccccccccr@{}}
    \toprule
    Method & \rotatebox{90}{Input} &\rotatebox{90}{mIoU (\%)} &\rotatebox{90}{others} &\rotatebox{90}{barrier} &\rotatebox{90}{bicycle} &\rotatebox{90}{bus} &\rotatebox{90}{car} &\rotatebox{90}{construction vehicle} &\rotatebox{90}{motorcycle} &\rotatebox{90}{pedestrian} &\rotatebox{90}{traffic corn} &\rotatebox{90}{trailer} &\rotatebox{90}{truck} &\rotatebox{90}{driveable area} &\rotatebox{90}{other flat} &\rotatebox{90}{sidewalk} &\rotatebox{90}{terrian} &\rotatebox{90}{mamade} &\multicolumn{1}{c}{\rotatebox{90}{vegetation}}
         \\ \midrule
    \multicolumn{20}{@{}l}{\emph{Zero-shot metric depth}} \\ \midrule   
    Metric3D~\cite{yin2023metric3d} &C &7.14 &0.00  &2.19 &3.16 &7.89 &6.42 &0.01 &4.93 &3.71 &0.12 &0.01 &6.88 &40.78 &0.18 &15.57 &23.24 &2.84 &3.43 \\
    DepthPro~\cite{bochkovskii2024depth} &C &7.40 &0.00  &4.44 &7.19 &14.87 &12.25 &0.01 &9.31 &8.16 &0.01 &0.01 &9.08 &27.83 &0.03 &10.25 &12.35 &3.32 &7.03 \\ \midrule
    \multicolumn{20}{@{}l}{\emph{Supervised with different signals}} \\ \midrule
    MonoScene~\cite{cao2022monoscene} & 3D & 6.06  & 1.75 & 7.23 & 4.26 & 4.93 & 9.38 & 5.67 & 3.98 & 3.01 & 5.90 & 4.45 & 7.17 & 14.91 & 6.32 & 7.92 & 7.43 & 1.01 & 7.65 \\
    OccFormer~\cite{zhang2023occformer} & 3D & 21.93 & 5.94 & 30.29 & 12.32 & 34.40 & 39.17 & 14.44 & 16.45 & 17.22 & 9.27 & 13.90 & 26.36 & 50.99 & 30.96 & 34.66 & 22.73 & 6.76 & 6.97 \\
    BEVFormer~\cite{li2022bevformer} & 3D  & 26.88 & 5.85 & 37.83 & 17.87 & 40.44 & 42.43 & 7.36 & 23.88 & 21.81 & 20.98 & 22.38 & 30.70 & {55.35} & 28.36 & 36.0 & 28.06 & 20.04 & 17.69 \\
    CTF-Occ~\cite{tian2024occ3d} & 3D & 28.53 & 8.09 & 39.33 & 20.56 & 38.29 & 42.24 & 16.93 & 24.52 & 22.72 & 21.05 & 22.98 & 31.11 & 53.33 & 33.84 & 37.98 & 33.23 & 20.79 & 18.00 \\
    TPVFormer~\cite{huang2023tri} & 3D  & 27.83 & 7.22 & 38.90 & 13.67 & 40.78 & 45.90 & 17.23 & 19.99 & 18.85 & 14.30 & 26.69 & 34.17 & 55.65 & 35.47 & 37.55 & 30.70 & 19.40 & 16.78 \\
    TPVFormer~\cite{huang2023tri} & L & 13.57 & 0.00 & 14.80 & 9.36 & 21.27 & 16.81 & 14.45 & 13.76 & 11.23 & 5.32 & 16.05 & 19.73 & 10.75 & 9.43 & 9.50 & 11.16 & 16.51 & 17.04 \\
    BEVDet~\cite{huang2021bevdet} &C &11.73 &2.09 &15.29 &0.00 &4.18 &12.97 &1.35 &0.00 &0.43 &0.13 &6.59 &6.66 &52.72 &19.04 &26.45 &21.78 &14.51 &15.26\\ \midrule
    \multicolumn{20}{@{}l}{\emph{Label free}} \\ \midrule
    SelfOcc (BEV)~\cite{huang2024selfocc} & C & 6.76 & 0.00 & 0.00  & 0.00  & 0.00  & 9.82  & 0.00  & 0.00 & 0.00  & 0.00  & 0.00  & 6.97  & 47.03  & 0.00  & 18.75  & 16.58  & 11.93  & 3.81  \\
    SelfOcc (TPV)~\cite{huang2024selfocc} & C & 9.30 & 0.00 & 0.15 & 0.66 & 5.46 &12.54 & 0.00 & 0.80 & 2.10 & 0.00 & 0.00 & 8.25 & \bf{55.49} & 0.00 & \bf{26.30} & \bf{26.54} & \bf{14.22} & 5.60 \\
    Ours (w/o training) &C & 7.04 &0.00 &3.91 &1.77 &7.15 &6.97 &0.00 &3.16 &1.17 &0.00 &0.00 &6.78 &47.37 &0.00 &16.24 &12.77 &6.21 &\bf{6.12} \\
    Ours &C &\bf{10.10} &0.00 &\bf{4.06} &\bf{6.26} &\bf{18.84} &\bf{16.87} &0.00 &\bf{8.55} &\bf{6.81} &\bf{0.01} &\bf{0.02} &\bf{12.16} &52.08 &\bf{0.03} &17.12 &19.28 &5.67 &3.86\\
    \bottomrule
    \end{tabular}
    }
    \label{tab:nusc}
\end{table*}
\section{Experiments}
\label{sec:exp}
\subsection{Experimantal Setting}
\noindent \textbf{Benchmark.} 
We evaluate our framework on two large-scale datasets for voxel occupancy prediction. 
The nuScenes dataset~\cite{caesar2020nuscenes} includes 28,130 frames for training and 6,019 for validation with 16 categories. Each frame captures six camera views surrounding the ego-vehicle. The image size is down-sampled from 900$\times$1600 to 512$\times$1408 for the main experiments and to 256$\times$704 for the ablation study. We report the mean Intersection-over-Union (mIoU) of the final checkpoint in nuScenes. For SemanticKITTI~\cite{behley2019semantickitti}, the dataset provides 3,834 pairs of binocular images for training and 815 for validation, covering 19 semantic categories. Consistent with common practice, we use only the left images for implementation. We employ class-agnostic IoU, precision, and recall to measure model performance, reporting the averaged scores over three runs.

\vspace{-1em}
\mypara{Zero-shot image primitives.}
For pixel semantics, we use CLIP~\cite{brown2020language,li2022languagedriven} to generate semantic maps over the plain images. In practice, $c_{\varnothing}$ (\cref{eq:sem}) is defined as the sky, as it falls outside the scope of interest in driving scenes. 
To reduce preparation time, the resolution of semantic map in nuScenes is downsampled to 450$\times$800. 
For monocular depth, we use DepthAnything~\cite{yang2024depth} as the source of relative depth. For a target view, we apply~\cref{eq:rec} twice on subsequent source views to determine the scene scale. We then fine-tune the pixel-wise scale and scene offset using AdamW~\cite{Loshchilov2019DecoupledWD} with 5,000 iterations and a learning rate of 1e-5. Again, to manage computational load, we downsample the depth map in nuScenes to 225$\times$400.

\mypara{Implementation.} 
We employ BEVFormer~\cite{li2022bevformer,huang2024selfocc} as the image encoder, following a two-layer MLP as task head. For nuScenes, we train the model over 5 epochs with an initial learning rate of 1e-4, decaying at epochs 1 and 3 by a factor of 0.5. The model is trained on 8 GPUs with 4 samples per GPU, taking approximately 10 hours. For SemanticKITTI, the initial learning rate is set to 2e-4, using a cosine scheduler over 10 epochs. This model is trained with a batch size of 1, which takes roughly 11 hours.

\subsection{Occupancy Prediction}
\label{sec:exp_occ}
\noindent \textbf{Validating the zero-shot label generation.}
For insight into the quality of our assembled supervision signals, we apply the pipeline of zero-shot label generation on nuScenes validation set and find it achieves a reasonable 7.04\% mIoU (penultimate row in~\cref{tab:nusc}), which outperforms learning-based MonoScene~\cite{cao2022monoscene} and SelfOcc(BEV)~\cite{huang2024selfocc}.

\paragraph{Comparison to zero-shot metric depth methods.} 
\cref{tab:nusc} shows our method fairly exceeds Metric3D~\cite{yin2023metric3d} and DepthPro~\cite{bochkovskii2024depth} in nuScenes by a clear margin of 2.96\% and 2.7\% mIoU, respectively. Specifically, DepthPro is superior in foreground (\eg, bus) while Metric3D excels in background (\eg, driveable area). 
The results evidence that existing zero-shot metric depth is incapable of occupancy prediction, underscoring the significance of our method.

\paragraph{Comparison to state-of-the-art on nuScenes.}
We compare our approach with current state-of-the-art methods. As shown in Tab.\ref{tab:nusc}, when no labels are available, our method outperforms SelfOcc~\cite{huang2024selfocc} by 3.34\% mIoU in BEV and 0.8\% mIoU in TPV on nuScenes. Generally, the proposed method exhibits superior performance, particularly for dynamic objects, which are central to driving scenarios. For example, our method achieves 18.84\% IoU for buses and 8.55\% IoU for motorcycles, surpassing SelfOcc (TPV) by absolute margins of 13.38\% and 7.75\% IoU, respectively. In terms of BEV representation, SelfOcc fails to predict 6 of the 8 dynamic object categories, except for cars and trucks.
Compared to supervised approaches, the proposed method is competitive with BEVDet~\cite{huang2021bevdet} (10.10\% \vs 11.73\% mIoU) and even surpasses MonoScene~\cite{cao2022monoscene} (10.10\% \vs 6.06\% mIoU). Notably, our method shows strength in surface prediction (\eg, 52.08\% vs. 10.75\% in the drivable area) compared to the LiDAR-guided TPVFormer~\cite{huang2023tri}, which struggles with surface predictions due to the sparsity of LiDAR signals, highlighting the advantages of a vision-centric paradigm. 

\begin{table}[t]
    \caption{Performance in SemantiKITTI. 
    $^*$ uses the depth information derived from MonoDepthv2~\cite{godard2019digging}. 
    We compare our method to the existing methods supervised by different signals.
    We use the class-agnostic IoU, precision, and recall (\%) to evaluate the geometry capability of the models. }
    \vspace{0mm}
    \centering
    \resizebox{1.0\columnwidth}{!}
    {
    \tablestyle{1pt}{1.05}
    \begin{tabular}{@{}l|ccc|cccc@{}}
    \toprule
         \multirow{2}{*}{Method} &\multicolumn{3}{c|}{Supervision} &\multirow{2}{*}{IoU (\%)}&\multirow{2}{*}{Precision (\%)} &\multirow{2}{*}{Recall (\%)}\\
          &3D &Depth &Image \\ \midrule
          \multicolumn{7}{@{}l}{\emph{Supervised with different signals}} \\ \midrule
          MonoScene~\cite{cao2022monoscene} &\cmark & & &37.14 &49.90 &59.24 \\ 
          LMSCNet$^{\rm rgb}$~\cite{roldao2020lmscnet} & &\cmark & &12.08 &13.00 &63.16 \\
          3DSketch$^{\rm rgb}$~\cite{chen20203d} & &\cmark & &12.01 &12.95 &62.31 \\
          AICNet$^{\rm rgb}$~\cite{li2020anisotropic} & &\cmark & &11.28 &11.84 &70.89\\
          MonoScene~\cite{cao2022monoscene} & &\cmark & &13.53 &16.98 &40.06\\
          MonoScene$^{*}$~\cite{cao2022monoscene} & & &\cmark &11.18 &13.15 &40.22\\ \midrule
          \multicolumn{7}{@{}l}{\emph{Label free}} \\ \midrule          
          SceneRF~\cite{cao2023scenerf} & & &\cmark &13.84 &17.28 &40.96\\
          SelfOcc (BEV)~\cite{huang2024selfocc} & & &\cmark &20.95 &\bf{37.27} &32.37\\
          SelfOcc (TPV)~\cite{huang2024selfocc} & & &\cmark &21.97 &34.83 &37.31\\
          Ours & & &\cmark &\bf{23.49$\pm$0.4} &35.15$\pm$0.5 &\bf{42.43$\pm$0.9} \\
         \bottomrule
    \end{tabular}
    }
    \label{tab:kitti_iou}
\end{table}

\mypara{Results on SemanticKITTI.}
For comparison, we report results from SceneRF~\cite{cao2023scenerf} and SelfOcc~\cite{huang2024selfocc}. Our method outperforms both SelfOcc and SceneRF across IoU and recall metrics by a significant margin. SceneRF demonstrates competitive recall but lower precision, while SelfOcc (BEV) achieves relatively higher precision but lower recall. In contrast, our method balances both precision and recall, leading to superior IoU, which reflects its strong capability in geometry representation learning.
We also provide the result of mIoU metric for reference in~\cref{tab:kitti_miou} of~\cref{supp:more_exp}.

\begin{figure}[t!]
\centering
\includegraphics[width=0.95\linewidth]{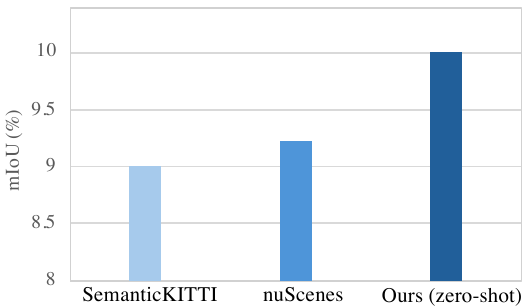}
\vspace{0mm}
\caption{
Comparison to fine-tuning scheme.
We fine-tune DepthAnything~\cite{yang2024depth} on in-domain (nuScenes~\cite{caesar2020nuscenes}) and out-of-domain (SemanticKITTI~\cite{behley2019semantickitti}) depth sources to obtain metric depth. To evaluate the quality of metric depth, we use them to train an occupancy network on nuScenes. Our zero-shot method can surpass them by a clear margin.}
\vspace{-3mm}
\label{fig:dataset}
\end{figure}
\mypara{Comparison to fine-tuning scheme.}
We further demonstrate the zero-shot capability of our method in metric depth estimation by comparing it with a fine-tuning approach. To evaluate depth quality, we fine-tune the depth VFM~\cite{yang2024depth} to obtain metric depth and use it, alongside the same zero-shot semantics, to train the occupancy network on nuScenes. Specifically, we fine-tune the VFM on in-domain (\ie, nuScenes) and out-of-domain (\ie, SemanticKITTI) depth sources, as illustrated in~\cref{fig:dataset}. Firstly, we fine-tune the VFM on SemanticKITTI. Due to the domain gap, fine-tuning on SemanticKITTI produces inaccurate metric depth for nuScenes, resulting in the lowest performance (9.01\% mIoU). Since nuScenes lacks a depth dataset, we project its point cloud to obtain pixel-wise depth and use this to fine-tune the VFM. However, we observe that this in-domain fine-tuning is inferior to our zero-shot method by 0.87\% mIoU, primarily due to the limitations of the training data quality. Specifically, only 1\% image pixels in nuScenes have the projected depth information due to the sparsity of LiDAR signals. In contrast, our method leverages the effective relative depth of VFM and adapts it to metric depth via novel view synthesis, without requiring ground-truth depth.

\mypara{Beyond pixel-space formulation.} 
The proposed method is formulated by NVS in the pixel space. Notably, our method is not exclusive to the 3D space formulation. We conduct an exploratory experiment integrating SLAM~\cite{teed2021droid} to enhance multi-frame consistency. \cref{fig:slam} of the Appendix shows SLAM can refine the geometry of the generated pseudo depth but tends to sparsify the signals, resulting in no significant improvement (\cref{tab:slam} of~\cref{supp:more_exp}). We will explore a combination of the signal density offered by NVS and the geometric precision of 3D space formulation in the future work.

\begin{table*}[t!]
     \caption{Ablation study on the metric depth adaptation, image backbone, image resolution, and the semantic map. The mIoU (\%) metric in nuScenes is reported. The default settings are marked as gray. SS: scene scale; PS: pixel scale; MM: mask moving object. }
    \begin{subtable}[h]{0.45\textwidth}
        \centering
    \caption{Image resolution}
    \begin{tabular}{ccc}
    \toprule
          Size&Mem. bound & mIoU \\ \midrule
         256$\times$704  &4868MB &9.83 \\
         384$\times$1056 &7845MB &9.93 \\
         \rowcolor{mygray} 512$\times$1408 &11288MB &10.10 \\
         896$\times$1600 &27799MB &10.19 \\
    \bottomrule
    \end{tabular}
        \label{tab:image_size}
     \end{subtable}
    \hfill
    \begin{subtable}[h]{0.5\textwidth}
        \centering
    \caption{Image backbone}
    \begin{tabular}{l cc}
    \toprule
        Backbone&Training cost &mIoU \\ \midrule
         \rowcolor{mygray} ResNet50 &10h &9.83\\
         ResNet101 &12h &9.92\\
         ResNet101+DCN &12.5h &9.95\\
    \bottomrule
    \end{tabular}
        \label{tab:image_backbone}
     \end{subtable}

    \hfill
    \begin{subtable}[h]{0.34\textwidth}
        \centering
        \caption{Resolution of depth map}
        \begin{tabular}{ccc }
        \toprule
         Size &s/image &mIoU \\ \midrule
        112$\times$200 &0.9 &9.44 \\
        \rowcolor{mygray}225$\times$400 &2.0 &9.83\\
        450$\times$800 &7.2 &9.85\\
        \bottomrule
       \end{tabular}
       \label{tab:ratio_dataset}
    \end{subtable}
    \hfill
    \begin{subtable}[h]{0.29\textwidth}
        \centering
        \caption{Depth components.}
    \begin{tabular}{cccc}
    \toprule
         SS &PS &MM& mIoU (\%) \\ \midrule
          & & & N/A
          \\
         \cmark & & &8.82 \\
         \cmark &\cmark & &9.34 \\
         \rowcolor{mygray}\cmark &\cmark &\cmark &9.83 \\
    \bottomrule
       \end{tabular}
       \label{tab:mfc}
    \end{subtable}
    \hfill
    \begin{subtable}[h]{0.35\textwidth}
        \centering
    \caption{Resolution of semantic map}
    \begin{tabular}{ccc}
    \toprule
         Size& Prep. time &mIoU \\ \midrule
         112$\times$200 &3d &9.02\\
         225$\times$400 &7d &9.39\\
         \rowcolor{mygray}450$\times$800 &15d &9.83\\
    \bottomrule
    \end{tabular}
        \label{tab:sem_size}
     \end{subtable}
\label{tab:other_ablation}
\end{table*}

\subsection{Ablation Study}
\begin{figure}[t!]
\centering
\includegraphics[width=0.95\linewidth]{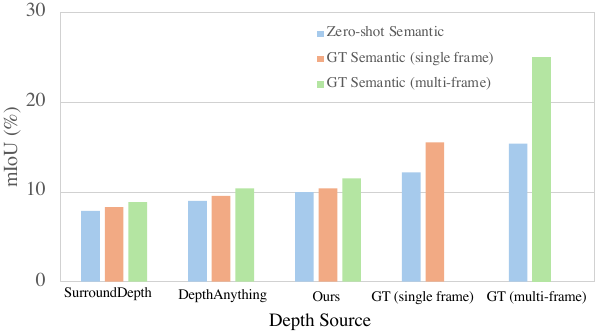}
\vspace{-2mm}
\caption{
Sensitivity and Extensibility.
We alternate the image semantics and depth to inspect the sensitivity and extensibility. We group the methods by their depth sources and apply different semantic sources within each group.
}
\vspace{0mm}
\label{fig:sem_depth}
\end{figure}

\paragraph{Semantic \vs Geometry.}
We examine the sensitivity and extensibility of our approach by varying the data sources for semantics and depth, as shown in~\cref{fig:sem_depth}. The $x$-axis represents depth sources, while the bar colours indicate semantic sources. Compared to SurroundDepth~\cite{wei2023surrounddepth} and DepthAnything~\cite{yang2024depth}, which rely on in-domain depth training, our zero-shot metric depth consistently outperforms them when using the same semantic source. Even against ground-truth (GT) depth (single frame), our method demonstrates competitive performance (10.10\% vs. 12.16\%) when using the same zero-shot semantics (blue bar).

By formulating 3D signals as an ensemble of \emph{independent} image primitives, our paradigm exhibits strong extensibility. For instance, under zero-shot depth, replacing the zero-shot semantics with GT semantics (multi-frame concatenation) improves performance from 10.10\% to 11.52\% mIoU. This highlights that the system can be augmented and extended by enhancing any individual image primitive.

\mypara{Ablating experimental settings.} 
We conduct a comprehensive ablation study on the experimental settings, summarised in Tab.~\ref{tab:other_ablation}, where the settings used in this work are highlighted in grey. In Tab.~\ref{tab:image_size}, increasing the image resolution from 256$\times$704 to 512$\times$1408 significantly improves performance while maintaining an acceptable memory footprint. For the image backbone, Tab.~\ref{tab:image_backbone} shows that ResNet50 achieves competitive results with a reasonable computational cost. Tab.~\ref{tab:ratio_dataset} examines the resolution of depth maps, where 225$\times$400 offers a good balance between performance and computation cost. The importance of the proposed technique for adapting metric depth (\cref{sec:scale}) is demonstrated in Tab.~\ref{tab:mfc}. Notably, without the proposed scene scale, naive application of pixel-wise scaling fails to calibrate metric depth effectively. As shown in Tab.~\ref{tab:sem_size}, we store semantic maps at a resolution of 450$\times$800, balancing performance and time efficiency. Additionally, we pre-train the network using coarse-grained annotations, when available, to investigate subsequent improvements, as discussed in~\cref{tab:pretrain} of the Appendix.

\vspace{2mm}

\section{Conclusion}
This work introduces a vision-centric framework for 3D occupancy prediction without requiring any 3D annotations. Specifically, we decouple 3D signals into two image primitives, semantic and depth information, bridging 2D VFMs with the 3D task. While existing VFMs for depth estimation provide promising relative depth, they are unsuitable for metric depth. To address this, we propose a coarse-to-fine strategy to calibrate relative depth into metric depth using novel view synthesis. Experiments on nuScenes and SemanticKITTI validate the effectiveness, comprehensive ablation studies further demonstrating the impact of each individual component.

\noindent \textbf{Limitation.}
In this work, we use the official category names as the vocabulary table to query semantic maps from CLIP~\cite{radford2021learning,li2022languagedriven}. However, label names can sometimes be ambiguous, leading to less accurate semantic maps than expected. For example, in nuScenes, labels such as \texttt{others} and \texttt{other flat} may cause confusion in CLIP due to their ambiguity, resulting in lower performance in these categories compared to fully-supervised methods. We leave addressing this limitation to future work.

\clearpage
{
    \small
    \bibliographystyle{ieeenat_fullname}
    \bibliography{egbib}
}

\clearpage
\onecolumn
\setcounter{page}{1}

\begin{center}
{\huge Supplementary Material} 
\end{center}

\section{Code Asset}

\mypara{Acknowledgement.} We detail the used benchmark in the main text. Our codebase is developed upon previous open-sourcing projects in ~\cref{tab:code}.  The authors acknowledge their works. 

\begin{table}[h]
    \caption{Acknowledgement to the used code asset.}
    \centering
    \resizebox{1.0\columnwidth}{!}
    {
    \tablestyle{20pt}{1.0}
    \begin{tabular}{@{}l|l|c|c}
    \toprule
         Experiments & \multicolumn{1}{c|}{URL}&Version&Licence \\ \midrule
         \multirow{2}{*}{Semantic VFM} &\url{https://github.com/isl-org/lang-seg} &707f7b7 &MIT\\ 
         &\url{https://github.com/openai/CLIP} &a1d0717 &MIT \\
         \midrule
         \multirow{3}{*}{Depth VFM} &\url{https://github.com/LiheYoung/Depth-Anything} &1e1c8d3 &Apache-2.0 \\
         &\url{https://github.com/DepthAnything/Depth-Anything-V2} &31dc977 &Apache-2.0 \\
         &\url{https://github.com/isl-org/ZoeDepth} &edb6daf &MIT\\
         \midrule
         \multirow{3}{*}{nuScenes} &\url{https://github.com/fundamentalvision/BEVFormer} &66b65f3 &Apache-2.0 \\
         &\url{https://github.com/Megvii-BaseDetection/BEVStereo} &62711e7 &MIT\\
         &\url{https://github.com/Tsinghua-MARS-Lab/Occ3D} &1c46609 &MIT\\
         \midrule
         \multirow{2}{*}{SemanticKITTI} &\url{https://github.com/huang-yh/SelfOcc} &7c22474 &MIT\\
         &\url{https://github.com/mrharicot/monodepth} &b76bee4 &UCLB ACP-A\\
    \bottomrule
    \end{tabular}
    }
    \label{tab:code}
\end{table}

\section{Extended Experiments}
\label{supp:more_exp}
\paragraph{Semantic occupancy prediction in SemanticKITTI.} 
In the main paper, we use the class-agnostic IoU metric to evaluate the geometry ability of our model and find it can exceed previous work~\cite{huang2024selfocc,cao2023scenerf} by a clear margin (\cref{tab:kitti_iou}). Here we provide the mIoU metric for reference in Tab.~\ref{tab:kitti_miou} and compare our model to some fully-supervised methods. We report the results with 3 volume settings: $12.8\times 12.8 \times 6.4$m$^3$, $25.6 \times 25.6 \times 6.4$m$^3$, and $51.2\times 51.2 \times 6.4$m$^3$.
Overall, our method exhibits competitive geometry ability as it can reach about 60\% ability of MonoScene~\cite{cao2022monoscene} using IoU metric.
In terms of semantic ability, it can approach about half of the mIoU of the fully-supervised MonoScene in two settings 12.8m and 25.6m. As an autonomous driving application, our method demonstrates reasonable performance. For instance, while LMSCNet~\cite{roldao2020lmscnet} fails to capture person and traffic signs in driving scenes, our model can provide useful information. Due to the semantic ambiguity, it is less effective for two categories, \texttt{other vehicle} and \texttt{other ground}, in generating the zero-shot semantic map. As a result, our method is less capable of these two categories.
It's worth noting that our method does not need any 3D annotations but label-free plain images, evidencing its zero-shot ability.

\begin{figure*}
    \centering
    \includegraphics[width=0.95\linewidth]{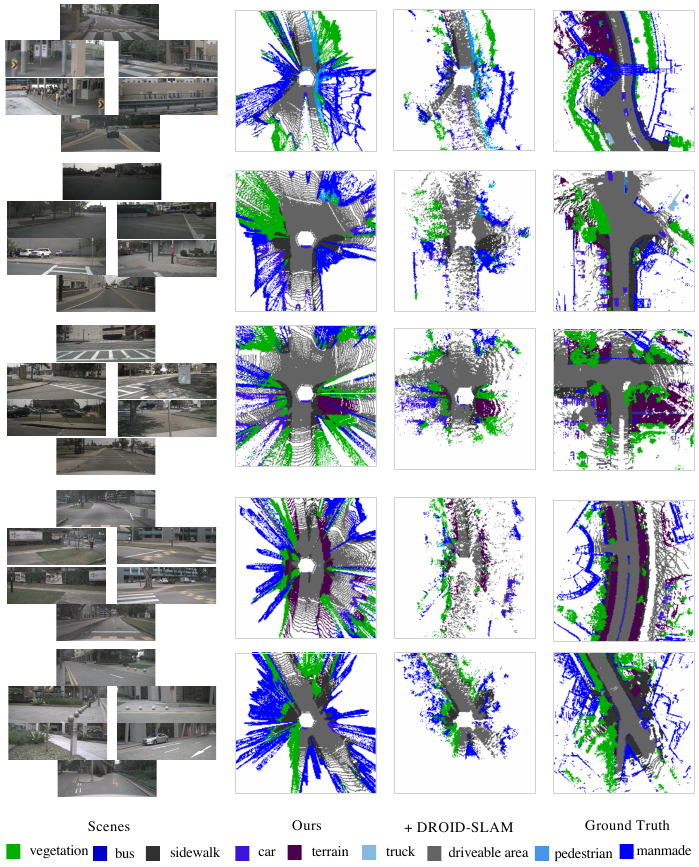}
    \caption{BEV visualization of the semantic voxel volume. We visualize the voxel volume constructed by our method (2nd col.). On top of that, we employ DROID-SLAM to impose the geometry constraint (3rd col.). We observe that SLAM-based method can enhance the geometry consistency but simultaneously reduce the supervision signals, \eg, driveable area.
    }
    \label{fig:slam}
\end{figure*}

\begin{table}[h!]
    \caption{Performance (\%) in SemantiKITTI. We compare the proposed method with two fully-supervised methods. Without any 3D annotations, our model exhibits competitive geometry ability (IoU) and reasonable semantic ability (mIoU).}
    \centering
    \resizebox{1.0\columnwidth}{!}
{\tablestyle{10pt}{1.0}
    \begin{tabular}{l|ccc|ccc|ccc}
    \toprule
         Method &\multicolumn{3}{c|}{MonoScene~\cite{cao2022monoscene}}& \multicolumn{3}{c|}{LMSCNet~\cite{roldao2020lmscnet}}& \multicolumn{3}{c}{Ours (zero-shot)} \\ \midrule
         Range&12.8m &25.6m &51.2m& 12.8m &25.6m &51.2m& 12.8m &25.6m &51.2m\\ \midrule
         IoU        &38.42 &38.55 &36.80 &65.52 &54.89 &38.36  &24.56 &26.12  &23.45\\
         Precision  &51.22 &51.96 &52.19 &86.51 &82.21 &77.60  &25.78 &29.93  &34.46\\
         Recall     &60.60 &59.91 &55.50 &72.98 &62.29 &43.13  &83.90 &67.23  &42.01\\ 
         \midrule
         mIoU       &12.25 &12.22 &11.30 &15.69 &14.13 &9.94   &6.23  &5.92   &4.31\\ 
         \midrule
         car        &24.34 &24.64 &23.29 &42.99 &35.41 &23.62  &24.76 &19.54  &12.16\\
         bicycle    &0.07 &0.23 &0.28 &0.00 &0.00 &0.00        &0.00  &0.00   &0.00\\
         motorcycle &0.05 &0.20 &0.59 &0.00 &0.00 &0.00        &0.35  &0.30   &0.23\\
         truck      &15.44 &13.84 &9.29 &0.76 &3.49 &1.69      &1.10  &0.67   &0.55\\
         other-veh. &1.18 &2.13 &2.63 &0.00 &0.00 &0.00        &0.00  &0.00   &0.00\\
         person     &0.90 &1.37 &2.00 &0.00 &0.00 &0.00        &0.29  &0.23   &0.18\\
         bicyclist  &0.54 &1.00 &1.07 &0.00 &0.00 &0.00        &0.00  &0.00   &0.00\\
         motorcyclist &0.00 &0.00 &0.00 &0.00 &0.00 &0.00      &0.03  &0.01   &0.01\\
         road       &57.37 &57.11 &55.89 &73.85 &67.56 &54.90  &40.58 &39.62  &32.84\\
         parking    &20.04 &18.60 &14.75 &15.63 &13.22 &9.89   &1.24  &0.78   &0.33\\
         sidewalk   &27.81 &27.58 &26.50 &42.29 &34.20 &25.43  &15.37 &11.47  &8.38\\
         other-ground &1.73 &2.00 &1.63 &0.00 &0.00 &0.00      &0.00  &0.00   &0.00\\
         building   &16.67 &15.97 &13.55 &22.46 &27.83 &14.55  &0.97  &4.01   &4.25\\
         fence      &7.57 &7.37 &6.60 &5.84 &4.42 &3.27        &4.65  &3.06   &1.83\\
         vegetation &19.52 &19.68 &17.98 &39.04 &33.32 &20.19  &7.39  &10.22  &7.42\\
         trunk      &2.02 &2.57 &2.44 &6.32 &3.01 &1.06        &0.15  &0.09   &0.04\\
         terrain    &31.72 &31.59 &29.84 &41.59 &41.51 &32.30  &17.89 &20.26  &11.95\\
         pole       &3.10  &3.79 &3.91 &7.28 &4.43 &2.04       &2.78  &1.55   &1.07\\
         traffic-sign &3.69 &2.54 &2.43 &0.00 &0.00 &0.00      &0.80  &0.72   &0.60\\

    \bottomrule
    \end{tabular}}
    \label{tab:kitti_miou}
\end{table}
\begin{table}[!t]
    \caption{Imposing geometry constraints to the monocular depth with DROID-SLAM.
    }
    \centering
    \resizebox{1.0\columnwidth}{!}
    {
    \tablestyle{2pt}{1.05}
    \begin{tabular}{@{}l|c|ccccccccccccccccr@{}}
    \toprule
    Method  &\rotatebox{90}{mIoU (\%)} &\rotatebox{90}{others} &\rotatebox{90}{barrier} &\rotatebox{90}{bicycle} &\rotatebox{90}{bus} &\rotatebox{90}{car} &\rotatebox{90}{construction vehicle} &\rotatebox{90}{motorcycle} &\rotatebox{90}{pedestrian} &\rotatebox{90}{traffic corn} &\rotatebox{90}{trailer} &\rotatebox{90}{truck} &\rotatebox{90}{driveable area} &\rotatebox{90}{other flat} &\rotatebox{90}{sidewalk} &\rotatebox{90}{terrian} &\rotatebox{90}{manmade} &\multicolumn{1}{c}{\rotatebox{90}{vegetation}}
         \\ \midrule
     Ours & 8.57  & 0.00 & 3.98 & 3.89 & 13.33 & 9.48 & \textbf{0.02} & 6.60 & \textbf{6.57} & \textbf{0.01} & 0.00 & \textbf{11.82} & \textbf{50.57} & \textbf{0.08} & 16.31 & \textbf{15.94} & 2.61 & \textbf{4.43} \\
    +DROID-SLAM~\cite{teed2021droid}& \textbf{8.69}  & 0.00 & \textbf{5.11} & \textbf{6.34} & \textbf{14.62} & \textbf{10.43} & 0.00 & \textbf{8.54} & 6.26 & \textbf{0.01} & \textbf{0.02} & 10.86 & 46.98 & \textbf{0.08} & \textbf{17.89} & 11.62 & \textbf{6.41} & 2.57 \\
    \bottomrule
    \end{tabular}
    }
    \label{tab:slam}
\end{table}
\paragraph{Can geometry constraints bring extra benefits?}
Noisy depth is a long-standing problem for monocular depth estimation as it lacks the geometry constraints in the output. We are interested in whether the refined depth can benefit the proposed method. We resort to SLAM to refine the depth by optimising the multi-view consistency. Specifically, the obtained RGB-D data (metric depth \& image) is the input to the DROID-SLAM~\cite{teed2021droid} for reconstructing the scene, generating the associated point cloud. We then project those point clouds to the image plane, obtaining the refined depth. As shown in Tab.~\ref{tab:slam}, the refined depth does not obviously improve the mIoU metric. In addition, DROID-SLAM can improve the performance of relatively static objects such as bicycle (6.34\% \vs 3.89\%) and manmade (6.41\% \vs 2.61\%). However, it is less favorable on surface prediction like driveable surface and terrain. We suspect that is due to a trade-off between the geometry consistency and density of the supervision. SLAM-based methods can enhance the geometry but simultaneously sparsifies the supervision signals (See the visualization Fig.~\ref{fig:slam}). 

\begin{table*}[ht]
    \centering
    \begin{minipage}{0.5\textwidth}
        \centering
        \caption{Improvement from other dimensions.}
    \begin{tabular}{@{}lc@{}}
    \toprule
         Method & mIoU \\ \midrule
         Ours &10.10\%\\
         +Depth pretraining &10.93\% \\
         +Obj. det. pretraining &11.16\% \\
    \bottomrule
    \end{tabular}
        \label{tab:pretrain}
    \end{minipage}
    \hspace{0.05\textwidth}
    \begin{minipage}{0.4\textwidth}
        \centering
        \caption{Comparison to SSL.}
    \begin{tabular}{@{}lcc}
    \toprule
         Method& \multicolumn{1}{c}{\emph{5\% lab.}} &\multicolumn{1}{c}{\emph{10\% lab.}} \\ \midrule
         Baseline &2.30\% &3.98\% \\
         SSL      &3.78\% &4.72\% \\
         SSL$^*$  &4.26\% &5.33\% \\
         Ours &10.62\%  &11.37\%\\ 
    \bottomrule
    \end{tabular}
        \label{tab:semi}
    \end{minipage}
\end{table*}
\paragraph{Prompting the model from other dimensions.}
A common trick for occupancy prediction is to pretrain the network with other 3D tasks. Since we presume no 3D labels are available, we do not use this trick in our main experiments for fair comparison. Given the difference in granularity, the labels for some \emph{coarse-grained} tasks (\eg, 3D object detection) are easier to obtain than that of voxel volume. We provide the additional experiments that promote the proposed method by pertaining the network on two coarse-grained tasks, 3D object detection and monocular depth estimation, for reader who may be interested. For object detection, we use the ground truth from nuScenes. For depth estimation, we use the calibrated metric depth from our model. In Tab.~\ref{tab:pretrain}, we find that the two auxiliary tasks can offer informative features that further facilitate the model. Possibly due to the noise, pretraining on pseudo-depth is less favourable.

\paragraph{Comparison to semi-supervised method.}
We decouple the 3D representation as the integral of 2D image primitives in~\cref{eq:decouple}.
One might argue that the alternative voxel volume $V$ could be derived from a pretrained occupancy network a.k.a. semi-supervised learning (SSL). Our response is that SSL is less flexible since it is limited to the predefined category space and would not generalise to distinct domains. For instance, the pretrained occupancy network in nuScenes cannot be applied to semanticKITTI due to the incompatibility of category space. In comparison, since we model $V$ as the ensemble of 2D image primitives, it is natural to utilize the zero-shot ability of 2D VFMs across various scenarios. We provide a quantitative comparison with the standard SSL method in Tab.~\ref{tab:semi}. Specifically, we presume that limited samples are annotated and use the labeled data to train an occupancy network. SSL will then produce pseudo-labels with the pretrained network and continuously train the network with both the pseudo-labels and limited labels. SSL$^*$ means producing the pseudo-labels in the second round. We observe that in two settings (5\% and 10\% labels are available), although SSL can improve the baseline, our method can consistently surpass SSL by a large margin ($\sim 6\%$).

\end{document}